# A Logic Programming Framework for Possibilistic Argumentation with Vague Knowledge

**Carlos I. Chesñevar**
Computer Science Dept.
University of Lleida
25001 Lleida, Spain
cic@eup.udl.es

**Guillermo R. Simari**
Computer Science Dept.
Universidad Nacional del Sur
8000 B.Blanca, Argentina
grs@cs.uns.edu.ar

**Teresa Alsinet**
Computer Science Dept.
University of Lleida
25001 Lleida, Spain
tracy@eup.udl.es

**Lluís Godo**
AI Research Institute
IIIA – CSIC
08193 Bellaterra, Spain
godo@iiia.csic.es

## Abstract

Defeasible argumentation frameworks have evolved to become a sound setting to formalize commonsense, qualitative reasoning from incomplete and potentially inconsistent knowledge. Defeasible Logic Programming (DeLP) is a defeasible argumentation formalism based on an extension of logic programming. Although DeLP has been successfully integrated in a number of different real-world applications, DeLP cannot deal with explicit uncertainty, nor with vague knowledge, as defeasibility is directly encoded in the object language. This paper introduces P-DeLP, a new logic programming language that extends original DeLP capabilities for qualitative reasoning by incorporating the treatment of possibilistic uncertainty and fuzzy knowledge. Such features will be formalized on the basis of PGL, a possibilistic logic based on Gödel fuzzy logic.

## 1 Introduction and motivations

In the last years defeasible argumentation frameworks have proven to be a successful approach to formalizing qualitative, commonsense reasoning from incomplete and potentially inconsistent knowledge [Chesñevar et al., 2000, Prakken and Vreeswijk, 2002]. As a consequence, argument-based frameworks have integrated in a number of real-world applications, such as automated text analysis [Hunter, 2001], intelligent web search [Chesñevar and Maguitman, 2004b], knowledge engineering [Carbogim et al., 2000] and clustering [Gómez and Chesñevar, 2004], among many others). Defeasible Logic Programming (or DeLP) [García and Simari, 2004] is one of such formalisms, combining results from defeasible argumentation theory and logic programming. Although DeLP has proven to be a suitable framework for building real-world applications that deal with incomplete and contradictory information in dynamic domains, it cannot deal with explicit uncertainty, nor with vague knowledge, as defeasible information is encoded in the object language using "defeasible rules".

This paper introduces P-DeLP, a new logic programming language that extends original DeLP capabilities for qualitative reasoning by incorporating the treatment of possibilistic uncertainty and fuzzy knowledge. Such features will be formalized on the basis of PGL [Alsinet and Godo, 2000, Alsinet and Godo, 2001], a possibilistic logic based on Gödel fuzzy logic. In PGL formulas are built over fuzzy propositional variables and the certainty degree of formulas is expressed with a necessity measure. In a logic programming setting, the proof method for PGL is based on a complete calculus for determining the maximum degree of possibilistic entailment of a fuzzy goal.

In the context of complex logic-programming frameworks (like the one provided by extended logic programming), PGL lacks of an adequate mechanism to handle contradictory information, as conflicting derivations can be found. In P-DeLP such conflicts will be solved using an argument-based inference engine. Formulas will be supported by arguments, which will have an attached necessity measure associated with the supported conclusion. The ultimate answer to queries will be given in terms of warranted arguments, computed through a dialectical analysis.

The rest of the paper is structured as follows. First, in Section 2 we formalize the syntax, semantics and the proof method of P-DeLP. In Section 3 we introduce the central notion of argument and a procedural mechanism for obtaining arguments. In Section 4 we formalize the notions of attack among arguments and the process of warrant in P-DeLP. Finally, in Section 5 we discuss related work and present the most impor-tant conclusions that have been obtained.

## 2 The P-DeLP programming language

As already pointed out our objective is to extend the DeLP programming language to deal with both vague knowledge and possibilistic uncertainty, we will refer to this extension as Possibilistic DeLP or P-DeLP. To this end, the base language of DeLP will be extended with fuzzy propositional variables and arguments will have an attached necessity measure associated with the supported conclusion. The ultimate answer to queries will be given in terms of warranted arguments, computed through a dialectical analysis.

The P-DeLP language $\mathcal{L}$ is defined from a set of fuzzy atoms (fuzzy propositional variables) $\{p, q, \ldots\}$ together with the connectives $\{\sim, \wedge, \leftarrow \}$. The symbol $\sim$ stands for *negation*. A *literal* $L \in \mathcal{L}$ is a ground (fuzzy) atom $q$ or a negated ground (fuzzy) atom $\sim q$, where $q$ is a (fuzzy) propositional variable. A *rule* in $\mathcal{L}$ is a formula of the form $Q \leftarrow L_1 \wedge \ldots \wedge L_n$, where $Q, L_1, \ldots, L_n$ are literals in $\mathcal{L}$. When $n = 0$, the formula $Q \leftarrow$ is called a *fact* and simply written as $Q$. In the following, capital and lower case letters will denote literals and atoms in $\mathcal{L}$, respectively.

On the one hand, fuzzy propositions provide us with a suitable representation model in situations where there is vague or imprecise information about the real world. For instance, the fuzzy statement "the engine speed is low" can be nicely represented by the fuzzy proposition $engine\_speed(low)$, where $low$ is a fuzzy set defined over the domain `revs_per_minute`, say an interval [0, 6000]. In the case $low$ actually denotes a crisp interval of number of revolutions, the above proposition is to be interpreted as "$\exists x \in low$ such that the engine speed is $x$". In the case $low$ denotes a fuzzy interval with a membership function $\mu_{low} : [0, 6000] \to [0, 1]$, the above proposition is interpreted in possibilistic terms as "for each $\alpha \in [0, 1]$, $\exists x \in [\mu_{low}]_\alpha$ such that the engine speed is $x$, is certain with a necessity of at least $1 - \alpha$", where $[\mu_{low}]_\alpha$ denotes the $\alpha$-cut of $\mu_{low}$, the set of values defined as $[\mu_{low}]_\alpha = \{u \in [0, 6000] \mid \mu_{low}(u) \geq \alpha\}$. So, fuzzy propositions can be seen as (flexible) restrictions on an existential quantifier [Dubois et al., 1998].

On the other hand, in this framework, negation is used to contradict statements represented by fuzzy propositions. For instance, in the case $low$ denotes a crisp interval of revolutions, $\sim speed(low)$ is interpreted as "$\neg[\exists x \in low$ such that the engine speed is $x]$", or equivalently "$\forall x \in low$, $x$ does not correspond with the engine speed".

A rigorous (and powerful) approach should be to define a first-order language with typed regular predicates and sorted fuzzy constants (cf. [Alsinet et al., 1999]) which would indeed represent, for instance, the fuzzy statement "the engine speed is low" as $engine\_speed(low)$ where $engine\_speed$ is a unary predicate of type (`revs_per_minute`) and $low$ is a fuzzy constant of sort `revs_per_minute`. In doing so, one is able to deal with partial matching between similar (fuzzy) constants, a very interesting feature. However, in this paper we are not considering yet the possibility of incorporating fuzzy unification between fuzzy constants in the language. Therefore, we will restrict ourselves to a simpler propositional language, where for instance the fuzzy statement $engine\_speed(low)$ will be simply represented as a fuzzy propositional variable $low\_speed$.

**Definition 1 (P-DeLP formulas)** *The set Wffs($\mathcal{L}$) of wffs in $\mathcal{L}$ are* facts *and* rules *built over the literals of $\mathcal{L}$. A certainty-weighted wff in $\mathcal{L}$ or* weighted clause *is a pair of the form $(\varphi, \alpha)$, where $\varphi$ is a wff in $\mathcal{L}$ and $\alpha \in [0, 1]$ expresses a lower bound for the certainty of $\varphi$ in terms of a necessity measure.*

The P-DeLP language is based on Possibilistic Gödel Logic or PGL [Alsinet and Godo, 2000]. There are three main reasons for choosing PGL as the underlying logic to model both uncertainty and fuzziness. First, we have proved that many-valued Gödel logic is fully compatible with an already proposed and suitable extension of necessity measures for fuzzy events, in the sense that Gödel logic allows us to define a well behaved and featured possibilistic semantics on top of it. Second, like in classical propositional logic programming systems, PGL enables us to define an efficient proof method by derivation based on a complete calculus for determining the maximum degree of possibilistic belief with which a fuzzy propositional variable can be entailed from a set of formulas. Finally, PGL can be extended with a partial matching mechanism between fuzzy propositional variables based on a necessity-like measure which preserves completeness for a particular class of formulas [Alsinet and Godo, 2001]. In our opinion, this is a key feature that justifies by itself the interest of such a logic programming system for defeasible argumentation under vague knowledge and possibilistic uncertainty.

The semantics of PGL [Alsinet and Godo, 2000] is given by *interpretations* $I$ of the fuzzy propositional variables into the real unit interval [0, 1] which are extended to wffs in $\mathcal{L}$ by means of the following rules:

$$I(L_1 \wedge \cdots \wedge L_n) = \min(I(L_1), \ldots, I(L_n))$$
$$I(Q \leftarrow \varphi) = \begin{cases} 1, & \text{if } I(\varphi) \leq I(Q) \\ I(Q), & \text{otherwise} \end{cases}$$

$$I(\sim q) = \begin{cases} 1, & \text{if } I(q) = 0 \\ 0, & \text{otherwise} \end{cases}$$

Certainty weights are employed to model statements of the form "$\varphi$ is $\alpha$-certain", where $\varphi$ represents vague knowledge about the real world. Within the possibilistic model of uncertainty, belief states are modelled by normalized possibility distributions $\pi : \mathcal{I} \to [0,1]$ on a set of interpretations $\mathcal{I}$. In our framework, the truth evaluation of a wff in $\mathcal{L}$ $\varphi$ in each interpretation $I$ is a value $I(\varphi) \in [0,1]$. Therefore, each formula does not induce a crisp set of interpretations, but a fuzzy set of interpretations $[\varphi]$, defining $\mu_{[\varphi]}(I) = I(\varphi)$, for each interpretation $I$. Hence, to measure the uncertainty induced on a formula by a possibility distribution on the set of interpretations $\mathcal{I}$ we have to consider some extension of the notion of necessity measure for fuzzy sets, in particular for fuzzy sets of interpretations. In [Dubois and Prade, 1991] the authors propose to define

$$N([\varphi] \mid \pi) = \inf_{I \in \mathcal{I}} \pi(I) \Rightarrow \mu_{[\varphi]}(I),$$

where $\mu_{[\varphi]}(I) = I(\varphi) \in [0,1]$ and $\Rightarrow$ is the reciprocal of Gödel's many-valued implication, which is defined as $x \Rightarrow y = 1$ if $x \leq y$ and $x \Rightarrow y = 1 - x$, otherwise.

Now let us go into formal definitions.

**Definition 2 (Possibilistic model)** *Let $\mathcal{I}$ be the set of many-valued, interpretations over the language $\mathcal{L}$. A possibilistic model is a normalized possibility distribution $\pi : \mathcal{I} \to [0,1]$ on the set of interpretations $\mathcal{I}$.*

A possibility distribution $\pi$ is normalized when there is at least one $I \in \mathcal{I}$ such that $\pi(I) = 1$. In other words, belief states modelled by normalized distributions are consistent states, in the sense that at least one interpretation (or state or possible world) has to be fully plausible.

**Definition 3 (Possibilistic entailment)** *A possibilistic model $\pi : \mathcal{I} \to [0,1]$ satisfies a clause $(\varphi, \alpha)$, written $\pi \models (\varphi, \alpha)$, iff $N([\varphi] \mid \pi) \geq \alpha$. Now let $\Gamma$ be a set of clauses in $\mathcal{L}$. We say that $\Gamma$ entails $(\varphi, \alpha)$, written $\Gamma \models (\varphi, \alpha)$, iff every possibilistic model satisfying all the clauses in $\Gamma$ also satisfies $(\varphi, \alpha)$.*

**Proposition 4** *Let $\Gamma$ be a set of clauses in $\mathcal{L}$. If $\Gamma$ is satisfiable, then $\Gamma \models \{(q, \alpha), (\sim q, \beta)\}$ iff either $\alpha = 0$ or $\beta = 0$.*[1]

In [Alsinet and Godo, 2000] we formalized a Hilbert-style axiomatization of PGL. Axioms of PGL are axioms of Gödel fuzzy logic weighted by 1 plus the triviality axiom $(\varphi, 0)$, and inference rules of PGL are a generalized modus ponens rule for necessity measures and a weight weakening rule. The proof method in PGL is defined for any certainty-weighted Gödel formula by deduction relative to the set of axioms and inference rules. In $\mathcal{L}$ wffs are either certainty-weighted facts or rules (with positive and negative literals) and the proof method should be oriented to goals (positive and negative literals). Then, for P-DeLP we will consider a simple and efficient calculus which will not need the whole axiomatization of PGL. But before we need to introduce some extra definitions and results.

---
[1]The proof is not included for space reasons.

**Definition 5 (Maximum degree of possibilistic entailment)** *The maximum degree of possibilistic entailment of a goal $Q$ from a set of clauses $\Gamma$, denoted by $\|Q\|_\Gamma$, is the greatest lower bound $\alpha \in [0,1]$ on the belief on $Q$ such that $\Gamma \models (Q, \alpha)$. Thus, $\|Q\|_\Gamma = \sup\{\alpha \in [0,1] \mid \Gamma \models (Q, \alpha)\}$.*

Follwing [Alsinet and Godo, 2000], one can prove that the *maximum degree of possibilistic entailment* of a goal $Q$ from a set of clauses $\Gamma$ is the *least necessity evaluation* of $Q$ given by the models of $\Gamma$.

To provide P-DeLP with a complete calculus for determining the maximum degree of possibilistic entailment we only need the triviality axiom of PGL and a particular instance of the generalized modus ponens rule of PGL:

**Axiom:** $(\varphi, 0)$

**Generalized modus ponens (GMP):**

$$\frac{(L_0 \leftarrow L_1 \wedge \cdots \wedge L_k, \gamma)}{(L_1, \beta_1), \ldots, (L_k, \beta_k)}$$
$$\overline{(L_0, \min(\gamma, \beta_1, \ldots, \beta_k))}$$

The GMP rule can be proven to be *sound* with respect to the many-valued and the possibilistic semantics of the underlying logic.

**Definition 6 (Degree of deduction)** *A goal $Q$ is deduced with a degree of deduction $\alpha$ from a set of clauses $\Gamma$, denoted $\Gamma \vdash (Q, \alpha)$, iff there exists a finite sequence of clauses $C_1, \ldots, C_m$ such that $C_m = (Q, \alpha)$ and, for each $i \in \{1, \ldots, m\}$, it holds that $C_i \in \Gamma$, $C_i$ is an instance of the axiom or $C_i$ is obtained by applying the above inference rule to previous clauses in the sequence.*

Due to the negation connective of P-DeLP, the GMP rule allows us to define a complete calculus for determining the maximum degree of possibilistic entailment of a goal from a set of clauses if we restrict ourselves to sets of clauses satisfying the following *forward reasoning* constraint: The possibilistic entailment degree of a goal $Q$ from a set of clauses $\Gamma$ must be univocally determined by those clauses of $Γ$ having $Q$ in their head or leading to one of these clauses by resolving them with other clauses by applying the GMP rule. The objective of the forward reasoning constraint is to ensure that, for any goal $Q$, $\|Q\|_\Gamma$ can be determined only from the subset $\Gamma_Q$ of clauses $(\varphi, \alpha)$ in $\Gamma$ for which either $Q$ is in the head of $\varphi$ or $Q$ *depends*[2] on the head of $\varphi$. Roughly speaking, with this requirement one wants to avoid having formulas of the form $(t \leftarrow p, 1)$ and $(\sim t, 1)$ together in $\Gamma$ since, due to the semantics of the negation connective, we would have that the clause $(\sim p, 1)$ should be derivable from $(t \leftarrow p, 1)$ and $(\sim t, 1)$, and thus, we should enable a kind of *modus tollens* inference mechanism. The forward reasoning constraint is ensured when for all literal $L$ appearing in the body of a rule, $\Gamma$ contains explicit information about $L$, i.e. either $(L, \alpha) \in \Gamma$ or $(L \leftarrow L_1 \wedge \cdots L_n, \alpha) \in \Gamma$ with $\alpha > 0$. A similar constraint was defined in [Alsinet and Godo, 2000], called there *context constraint*, for preserving completeness when extending PGL with a fuzzy unification mechanism between fuzzy constants. At this point we are ready to define the syntactic counterpart of maximum degree of possibilistic entailment.

**Definition 7 (Maximum degree of deduction)**
*The maximum degree of deduction of a goal $Q$ from a set of clauses $\Gamma$, denoted $|Q|_\Gamma$, is the greatest $\alpha \in [0, 1]$ such that $\Gamma \vdash (Q, \alpha)$.*

As the only inference rule of our proof method is the GMP rule within a logic programming framework in which $\Gamma$ is always a finite set of clauses, there exists a finite number of proofs of a goal $Q$ from $\Gamma$, and thus, the above definition turns into $|Q|_\Gamma = \max\{\alpha \in [0, 1] \mid \Gamma \vdash (Q, \alpha)\}$.

Finally, following [Alsinet and Godo, 2000, Alsinet and Godo, 2001], *completeness* for P-DeLP reads as follows: Let $\Gamma$ be a set of clauses satisfying the forward reasoning constraint and let $Q$ be a goal. Then, $\|Q\|_\Gamma = |Q|_\Gamma$. From now on, we will consider clauses in $\mathcal{L}$ satisfying the forward reasoning constraint.

## 3　Argumentation in P-DeLP

In the last section we formalized the many-valued and the possibilistic semantics of the underlying logic of P-DeLP. In this section we formalize the procedural mechanism for building arguments in P-DeLP.

We distinguish between *certain* and *uncertain* clauses. A clause $(\varphi, \alpha)$ will be referred as certain if $\alpha = 1$ and uncertain, otherwise. Moreover, a set of clauses $\Gamma$ will be deemed as *contradictory*, denoted $\Gamma \vdash \bot$, if $\Gamma \vdash (q, \alpha)$ and $\Gamma \vdash (\sim q, \beta)$, with $\alpha > 0$ and $\beta > 0$, for some atom $q$ in $\mathcal{L}$. Notice that if $\Gamma$ is a contradictory set of clauses for some atom $q$ in $\mathcal{L}$, $\Gamma$ is not satisfiable and there exist $\Gamma_1 \subset \Gamma$ and $\Gamma_2 \subset \Gamma$ such that $\Gamma_1$ and $\Gamma_2$ are satisfiable and $|q|_{\Gamma_1} > 0$ and $|\sim q|_{\Gamma_2} > 0$.

**Example 8** *Consider the set $\Gamma = \{ (p \leftarrow q, 0.5), (\sim p \leftarrow q \wedge r, 0.3), (q, 0.2), (r, 1) \}$. Then $\Gamma$ is contradictory, whereas $\Gamma \setminus \{(r, 1)\}$ is not.*

A P-DeLP program is a set of clauses in $\mathcal{L}$ in which we distinguish certain from uncertain information. As additional requirement, certain knowledge is required to be non-contradictory. Formally:

**Definition 9 (P-DeLP program)** *A P-DeLP program $\mathcal{P}$ (or just program $\mathcal{P}$) is a pair $(\Pi, \Delta)$, where $\Pi$ is a non-contradictory finite set of certain clauses, and $\Delta$ is a finite set of uncertain clauses.*

**Example 10** *Consider an intelligent agent controlling an engine with three switches $sw1$, $sw2$ and $sw3$. These switches regulate different features of the engine, such as pumping system, speed, etc. This agent may have the following certain and uncertain knowledge about how this engine works, e.g.: – If the pump is clogged, then the engine gets no fuel.*
*– When $sw1$ is on, normally fuel is pumped properly.*
*– When fuel is pumped properly, fuel seems to work ok.*
*– When $sw2$ is on, usually oil is pumped.*
*– When oil is pumped, usually it works ok.*
*– When there is oil and fuel, usually the engine works ok.*
*– When there is heat, then the engine is usually not ok.*
*– When there is heat, normally there are oil problems.*
*– When fuel is pumped and speed is low, then there are reasons to believe that the pump is clogged.*
*– When $sw2$ is on, usually speed is low.*
*– When $sw2$ and $sw3$ are on, usually speed is not low.*
*– When $sw3$ is on, usually fuel is ok.*
*Suppose also that the agent knows some particular facts: $sw1$, $sw2$ and $sw3$ are on, and there is heat. The knowledge of such an agent can be modelled by the program $\mathcal{P}_{engine}$ shown in Fig. 1. Note that uncertainty is assessed in terms of different necessity measures.*

Next we will introduce the notion of *argument* in P-DeLP. Informally, a argument $\mathcal{A}$ is a tentative proof (as it relies to some extent on uncertain, possibilistic information) supporting a given conclusion $Q$ with a necessity measure $\alpha$. Formally:

**Definition 11 (Argument. Subargument)**
*Given a P-DeLP program $\mathcal{P} = (\Pi, \Delta)$, a set $\mathcal{A} \subseteq \Delta$ of uncertain clauses is an* argument *for a goal $Q$ with necessity measure $\alpha > 0$ (denoted $\langle \mathcal{A}, Q, \alpha \rangle$) iff:*

---

[2] We say that $Q$ *depends* on $P$ in $\Gamma$ if $\Gamma$ contains a set of clauses $\{(\varphi_1, \alpha_1), \ldots, (\varphi_k, \alpha_k)\}$, with $k \geq 1$, such that $P$ appears in the body of $\varphi_1$, the head of $\varphi_k$ is $Q$, and the head of $\varphi_i$ appears in the body of $\varphi_{i+1}$, with $i \in \{1, \ldots, k-1\}$.

(1) $(\sim\!fuel\_ok \leftarrow pump\_clog, 1)$
(2) $(sw1, 1)$
(3) $(sw2, 1)$
(4) $(sw3, 1)$
(5) $(heat, 1)$
(6) $(pump\_fuel \leftarrow sw1, 0.6)$
(7) $(fuel\_ok \leftarrow pump\_fuel, 0.3)$
(8) $(pump\_oil \leftarrow sw2, 0.8)$
(9) $(oil\_ok \leftarrow pump\_oil, 0.8)$
(10) $(engine\_ok \leftarrow fuel\_ok \wedge oil\_ok, 0.3)$
(11) $(\sim\!engine\_ok \leftarrow heat, 0.95)$
(12) $(\sim\!oil\_ok \leftarrow heat, 0.9)$
(13) $(pump\_clog \leftarrow pump\_fuel \wedge low\_speed, 0.7)$
(14) $(low\_speed \leftarrow sw2, 0.8)$
(15) $(\sim\!low\_speed \leftarrow sw2, sw3, 0.8)$
(16) $(fuel\_ok \leftarrow sw3, 0.9)$

Figure 1: P-DeLP program $\mathcal{P}_{eng}$ (example 10)

*1)* $\Pi \cup \mathcal{A} \vdash (Q, \alpha)$;
*2)* $\Pi \cup \mathcal{A}$ *is non contradictory; and*
*3)* $\mathcal{A}$ *is minimal wrt set inclusion (i.e. there is no $\mathcal{A}_1 \subset \mathcal{A}$ which satisfies conditions (1) and (2)).*

Let $\langle \mathcal{A}, Q, \alpha \rangle$ and $\langle \mathcal{S}, R, \beta \rangle$ be two arguments. We will say that $\langle \mathcal{S}, R, \beta \rangle$ is a *subargument* of $\langle \mathcal{A}, Q, \alpha \rangle$ iff $\mathcal{S} \subseteq \mathcal{A}$. Notice that the goal $R$ may be a subgoal associated with the goal $Q$ in the argument $\mathcal{A}$.

Note that an argument must satisfy certain requirements. First, the conclusion $Q$ should follow via $\vdash$ from $\Pi \cup \mathcal{A}$. Second, it should not be the case that $\mathcal{A}$ together with $\Pi$ turns out to be contradictory. The third requirement operates as a kind of Occam's razor principle [García and Simari, 2004] on the uncertain information used for concluding $Q$. It must be remarked that the definition of argument satisfying these three requirements can be traced back to the Simari-Loui framework [Simari and Loui, 1992]. Given a program $\mathcal{P} = (\Pi, \Delta)$, we define the following procedural rules to construct arguments:

**1) Building arguments from facts (INTF)**

$$\frac{(Q, 1)}{\langle \emptyset, Q, 1 \rangle} \qquad \frac{(Q, \alpha),\ \Pi \cup \{(Q, \alpha)\} \not\vdash \bot,\ \text{with}\ \alpha < 1,}{\langle \{(Q, \alpha)\}, Q, \alpha \rangle}$$

for any weighted fact $(Q, \alpha) \in \mathcal{P}$

**3) Building Arguments by GMP (MPA):**

$$\frac{\langle \mathcal{A}_1, L_1, \alpha_1 \rangle \quad \langle \mathcal{A}_2, L_2, \alpha_2 \rangle \ \ldots\ \langle \mathcal{A}_k, L_k, \alpha_k \rangle}{\langle \bigcup_{i=1}^{k} \mathcal{A}_i \cup \{(L_0 \leftarrow L_1 \wedge L_2 \wedge \ldots \wedge L_k, \gamma)\}, L_0, \beta \rangle}$$

$(L_0 \leftarrow L_1 \wedge L_2 \wedge \ldots \wedge L_k, \gamma)$ with $\gamma < 1$
$\Pi \cup \{(L_0 \leftarrow L_1 \wedge L_2 \wedge \ldots \wedge L_k, \gamma)\} \cup \bigcup_{i=1}^{k} \mathcal{A}_i \not\vdash \bot$

for any weighted rule $(L_0 \leftarrow L_1 \wedge L_2 \wedge \ldots \wedge L_k, \gamma) \in \Delta$, with $\beta = min(\alpha_1, \ldots, \alpha_k, \gamma)$.

**4) Extending Arguments (EAR):**

$$\frac{\langle \mathcal{A}, P, \alpha \rangle \quad \Pi \cup \{(P, \alpha)\} \vdash (Q, \alpha)}{\langle \mathcal{A}, Q, \alpha \rangle}$$

for any argument $\langle \mathcal{A}, P, \alpha \rangle$, whenever $(Q, \alpha)$ follows from $\Pi \cup \{(P, \alpha)\}$.

The basic idea with the argument construction procedure is to keep a trace of the set $\mathcal{A}$ of all uncertain information used to derive a given goal $Q$ with necessity degree $\alpha$. Appropriate preconditions ensure that the proof obtained always follows Cond. 2 in Def. 11. Given a program $\mathcal{P}$, rule INTF allows to construct arguments from facts. An empty argument can be obtained for any certain fact in $\mathcal{P}$. An argument concluding an uncertain fact $(Q, \alpha)$ in $\mathcal{P}$ can be derived whenever assuming $(Q, \alpha)$ is not contradictory wrt the set $\Pi$ in $\mathcal{P}$. Rule MPA accounts for modus ponens with a uncertain rule $(L_0 \leftarrow L_1 \wedge L_2 \wedge \ldots \wedge L_k, \gamma)$, with $\gamma < 1$. Note that there must exist arguments $\langle \mathcal{A}_1, L_1, \alpha_1 \rangle, \ldots, \langle \mathcal{A}_1, L_k, \alpha_1 \rangle$ for every literal in the antecedent of the rule. MPA is applicable whenever no contradiction wrt $\Pi$ results when assuming the rule $(L_0 \leftarrow L_1 \wedge L_2 \wedge \ldots \wedge L_k, \gamma)$ and the sets $\mathcal{A}_1, \ldots, \mathcal{A}_k$ corresponding to the arguments associated with the antecedent of the rule. Finally, the rule EAR stands for extending a given argument $\langle \mathcal{A}, P, \alpha \rangle$ on the basis of certain knowledge. As any argument is non contradictory wrt $\Pi$, making new inferences from $\Pi \cup (P, \alpha)$ cannot lead to contradictions, and hence $\langle \mathcal{A}, Q, \alpha \rangle$ is also valid whenever $\Pi \cup \{(P, \alpha)\} \vdash (Q, \alpha)$. Note that in this case the necessity measure of $Q$ is the same as the one computed for $P$, as no uncertainty is involved.

**Example 12** *Consider the program $\mathcal{P}_{eng}$ in Example 10. An argument $\langle \mathcal{B}, fuel\_ok, 0.3 \rangle$ can be derived from $\mathcal{P}_{eng}$ as follows:*

*i)* $\langle \emptyset, sw1, 1 \rangle$ *from (2) via INTF.*
*ii)* $\langle \mathcal{B}', pump\_fuel, 0.6 \rangle$ *from (6) and i) via MPA.*
*iii)* $\langle \mathcal{B}, fuel\_ok, 0.3 \rangle$ *from (7) and ii) via MPA.*

*where* $\mathcal{B}'=\{(pump\_fuel \leftarrow sw1, 0.6)\}$ *and* $\mathcal{B}=\{(pump\_fuel \leftarrow sw1, 0.6); (fuel\_ok \leftarrow pump\_fuel, 0.3)\}$. *Similarly, an argument $\langle \mathcal{C}, oil\_ok, 0.8 \rangle$ can be derived from $\mathcal{P}_{eng}$ using the rules (3), (8) and (9) via INTC, MPA, and MPA, resp., with $\mathcal{C} = \{(pump\_oil \leftarrow sw2, 0.8); (oil\_ok \leftarrow pump\_oil, 0.8)\}$.[3] Finally, an argument $\langle \mathcal{A}_1, engine\_ok, 0.3 \rangle$ can be derived from $\mathcal{P}_{eng}$ as follows:*

*i)* $\langle \mathcal{B}, fuel\_ok, 0.3 \rangle$ *as shown above.*
*ii)* $\langle \mathcal{C}, oil\_ok, 0.8 \rangle$ *as shown above.*
*iii)* $\langle \mathcal{A}_1, engine\_ok, 0.3 \rangle$ *from i), ii), 10) via MPA.*

*where* $\mathcal{A}_1=\{(engine\_ok \leftarrow fuel\_ok \wedge oil\_ok, 0.3)\} \cup \mathcal{B} \cup \mathcal{C}$. *Note that $\langle \mathcal{C}, oil\_ok, 0.8 \rangle$ and $\langle \mathcal{B}, fuel\_ok, 0.3 \rangle$ are subarguments of $\langle \mathcal{A}_1, engine\_ok, 0.3 \rangle$.*

---

[3] For the sake of clarity, we use semicolons to separate elements in an argument $\mathcal{A} = \{e_1\ ;\ e_2\ ;\ \ldots;\ e_k\ \}$.

## 4 Computing Warrant in P-DeLP

### 4.1 Counterargumentation and defeat

Given a program $\mathcal{P}$, it can be the case that there exist conflicting arguments for complementary literals $\langle \mathcal{A}_1, q, \alpha_1 \rangle$ and $\langle \mathcal{A}_2, \sim q, \alpha_2 \rangle$. Such conflict among arguments will be formalized by the notions of counterargument and defeat presented next.

**Definition 13 (Counterargument)** Let $\mathcal{P}$ be a program, and let $\langle \mathcal{A}_1, Q_1, \alpha_1 \rangle$ and $\langle \mathcal{A}_2, Q_2, \alpha_2 \rangle$ be two arguments wrt $\mathcal{P}$. We will say that $\langle \mathcal{A}_1, Q_1, \alpha_1 \rangle$ counterargues $\langle \mathcal{A}_2, Q_2, \alpha_2 \rangle$ iff there exists a subargument (called disagreement subargument) $\langle \mathcal{S}, Q, \beta \rangle$ of $\langle \mathcal{A}_2, Q_2, \alpha_2 \rangle$ such that $\Pi \cup \{(Q_1, \alpha_1), (Q, \beta)\}$ is contradictory.

**Example 14** Consider $\langle \mathcal{A}_1, engine\_ok, 0.3 \rangle$ given in Example 12 wrt the program $\mathcal{P}_{eng}$. A counteragument for $\langle \mathcal{A}_1, engine\_ok, 0.3 \rangle$ can be found, namely $\langle \mathcal{A}_2, \sim fuel\_ok, 0.6 \rangle$, obtained from (2), (3), (6), (14), (13) and (1) by applying INTF, INTF, MPA, MPA, MPA, and EAR, resp., with $\mathcal{A}_2 = \{ (pump\_fuel \leftarrow sw1, 1), (low\_speed \leftarrow sw2, 1), (pump\_clog \leftarrow pump\_fuel \wedge low\_speed, 1)\}$. Argument $\langle \mathcal{A}_2, \sim fuel\_ok, 0.6 \rangle$ is a counterargument for $\langle \mathcal{A}_1, engine\_ok, 0.3 \rangle$ as there exists a subargument $\langle \mathcal{B}, fuel\_ok, 0.3 \rangle$ in $\langle \mathcal{A}_1, engine\_ok, 0.3 \rangle$ (see Example 12) such that $\Pi \cup \{(fuel\_ok, 0.3), (\sim fuel\_ok, 0.6)\}$ is contradictory.

Defeat among arguments involves a *preference criterion* on conflicting arguments, defined on the basis of necessity measures associated with arguments.

**Definition 15 (Preference criterion $\succeq$)** Let $\mathcal{P}$ be a program, and and let $\langle \mathcal{A}_1, Q_1, \alpha_1 \rangle$ and $\langle \mathcal{A}_2, Q_2, \alpha_2 \rangle$ be conflicting arguments in $\mathcal{P}$. We will say that $\langle \mathcal{A}_1, Q_1, \alpha_1 \rangle$ is preferred over $\langle \mathcal{A}_2, Q_2, \alpha_2 \rangle$ (denoted $\langle \mathcal{A}_1, Q_1, \alpha_1 \rangle \succeq \langle \mathcal{A}_2, Q_2, \alpha_2 \rangle$) iff $\alpha_1 \geq \alpha_2$.
If $\alpha_1 > \alpha_2$, then we will say that $\langle \mathcal{A}_1, Q_1, \alpha_1 \rangle$ is strictly preferred over $\langle \mathcal{A}_2, Q_2, \alpha_2 \rangle$. Otherwise, if $\alpha_1 = \alpha_2$ we will say that both arguments are equipreferred, denoted $\langle \mathcal{A}_2, Q_2, \alpha_2 \rangle) \approx \langle \mathcal{A}_1, Q_1, \alpha_1 \rangle$.

**Definition 16 (Defeat)** Let $\mathcal{P}$ be a program, and let $\langle \mathcal{A}_1, Q_1, \alpha_1 \rangle$ and $\langle \mathcal{A}_2, Q_2, \alpha_2 \rangle$ be two arguments in $\mathcal{P}$. We will say that $\langle \mathcal{A}_1, Q_1, \alpha_1 \rangle$ defeats $\langle \mathcal{A}_2, Q_2, \alpha_2 \rangle$ (or equivalently $\langle \mathcal{A}_1, Q_1, \alpha_1 \rangle$ is a defeater for $\langle \mathcal{A}_2, Q_2, \alpha_2 \rangle$) iff
1) $\langle \mathcal{A}_1, Q_1, \alpha_1 \rangle$ counterargues $\langle \mathcal{A}_2, Q_2, \alpha_2 \rangle$ with disagreement subargument $\langle \mathcal{A}, Q, \alpha \rangle$.
2) Either it holds that $\alpha_1 > \alpha$, in which case $\langle \mathcal{A}_1, Q_1, \alpha_1 \rangle$ will be called a proper defeater for $\langle \mathcal{A}_2, Q_2, \alpha_2 \rangle$, or $\alpha_1 = \alpha$, in which case $\langle \mathcal{A}_1, Q_1, \alpha_1 \rangle$ will be called a blocking defeater for $\langle \mathcal{A}_2, Q_2, \alpha_2 \rangle$.

**Example 17** Consider $\langle \mathcal{A}_1, engine\_ok, 0.3 \rangle$ and $\langle \mathcal{A}_2, \sim fuel\_ok, 0.6 \rangle$ in Example 14. Then $\langle \mathcal{A}_2, \sim fuel\_ok, 0.6 \rangle$ is a proper defeater for $\langle \mathcal{A}_1, engine\_ok, 0.3 \rangle$, as $\langle \mathcal{A}_2, \sim fuel\_ok, 0.6 \rangle$ counterargues $\langle \mathcal{A}_1, engine\_ok, 0.3 \rangle$ with disagreement subargument $\langle \mathcal{B}, fuel\_ok, 0.3 \rangle$, and $0.6 > 0.3$.

### 4.2 Computing Warrant via Dialectical Trees

Given an argument $\langle \mathcal{A}, Q, \alpha \rangle$, the definitions of counterargument and defeat allows to detect whether other possible arguments $\langle \mathcal{B}_1, Q_1, \alpha_1 \rangle \ldots \langle \mathcal{B}_k, Q_k, \alpha_k \rangle$ are defeaters for $\langle \mathcal{A}, Q, \alpha \rangle$. Should the argument $\langle \mathcal{A}, Q, \alpha \rangle$ be defeated, then it would be no longer supporting its conclusion $Q$. However, since defeaters are arguments, they may on their turn be defeated. That prompts for a complete recursive dialectical analysis to determine which arguments are ultimately defeated. Ultimately undefeated arguments will be marked as *U-nodes*, and the defeated ones as *D-nodes*. To characterize this process we will introduce some auxiliary notions.

An *argumentation line* starting in an argument $\langle \mathcal{A}_1, Q_1, \alpha_1 \rangle$ (denoted $\lambda^{\langle \mathcal{A}_1, Q_1, \alpha_1 \rangle}$) is a sequence $[\langle \mathcal{A}_0, Q_0, \alpha_0 \rangle, \langle \mathcal{A}_1, Q_1, \alpha_1 \rangle, \ldots, \langle \mathcal{A}_n, Q_n, \alpha_n \rangle, \ldots]$ that can be thought of as an exchange of arguments between two parties, a *proponent* (evenly-indexed arguments) and an *opponent* (oddly-indexed arguments). Each $\langle \mathcal{A}_i, Q_i, \alpha_i \rangle$ is a defeater for the previous argument $\langle \mathcal{A}_{i-1}, Q_{i-1}, \alpha_{i-1} \rangle$ in the sequence, $i > 0$. In order to avoid *fallacious* reasoning, argumentation theory imposes additional constraints on such an argument exchange to be considered rationally acceptable wrt a P-DeLP program $\mathcal{P}$, namely:

1) **Non-contradiction:** given an argumentation line $\lambda$, the set of arguments of the proponent (resp. opponent) should be *non-contradictory* wrt $\mathcal{P}$.
2) **No circular argumentation:** no argument $\langle \mathcal{A}_j, Q_j, \alpha_j \rangle$ in $\lambda$ is a sub-argument of an argument $\langle \mathcal{A}_i, Q_i, \alpha_i \rangle$ in $\lambda$, $i < j$.
3) **Progressive argumentation:** every blocking defeater $\langle \mathcal{A}_i, Q_i, \alpha_i \rangle$ in $\lambda$ is defeated by a proper defeater $\langle \mathcal{A}_{i+1}, Q_{i+1}, \alpha_{i+1} \rangle$ in $\lambda$.

The first condition disallows the use of contradictory information on either side (proponent or opponent). The second condition eliminates the "circular reasoning" fallacy. The last condition enforces the use of a stronger argument to defeat an argument which acts as a blocking defeater. An argumentation line satisfying the above restrictions is called *acceptable*, and can be proven to be finite.

**Example 18** *Consider $\langle \mathcal{A}_1, engine\_ok, 0.3\rangle$ and the associated defeater $\langle \mathcal{A}_2, \sim fuel\_ok, 0.6\rangle$ in Example 17. Note that $\langle \mathcal{A}_2, \sim fuel\_ok, 0.6\rangle$ has the associated subargument $\langle \mathcal{A}_2', low\_speed, 0.8\rangle$, with $\mathcal{A}_2' = \{(low\_speed \leftarrow sw2, 0.8)\}$. From the program $\mathcal{P}_{eng}$ (Fig. 1) a blocking defeater for $\langle \mathcal{A}_2, \sim fuel\_ok, 0.6\rangle$ can be derived, namely $\langle \mathcal{A}_3, \sim low\_speed, 0.8\rangle$, obtained from (3), (4) and (15) via INTF, INTF and MPA, resp. Note that this third defeater can be thought of as an answer of the proponent to the opponent, reinstating the first argument $\langle \mathcal{A}_1, engine\_ok, 0.3\rangle$, as it defeats the opponent's defeater $\langle \mathcal{A}_2, \sim fuel\_ok, 0.6\rangle$. The above situation can be expressed in the following argumentation line: $[\langle \mathcal{A}_1, engine\_ok, 0.3\rangle, \langle \mathcal{A}_2, \sim fuel\_ok, 0.6\rangle, \langle \mathcal{A}_3, \sim low\_speed, 0.8\rangle]$. Note that the proponent's last defeater in the above sequence could be on its turn defeated by a blocking defeater $\langle \mathcal{A}_2', low\_speed, 0.8\rangle$, resulting in $[\langle \mathcal{A}_1, engine\_ok, 0.3\rangle, \langle \mathcal{A}_2, \sim fuel\_ok, 0.6\rangle, \langle \mathcal{A}_3, \sim low\_speed, 0.8\rangle, \langle \mathcal{A}_2', low\_speed, 0.8\rangle \ldots]$. However, such line is* not acceptable, *as it violates the condition of non-circular argumentation.*

Given a program $\mathcal{P}$ and an argument $\langle \mathcal{A}_0, Q_0, \alpha_0\rangle$, the set of all acceptable argumentation lines starting in $\langle \mathcal{A}_0, Q_0, \alpha_0\rangle$ accounts for a whole dialectical analysis for $\langle \mathcal{A}_0, Q_0, \alpha_0\rangle$ (i.e. all possible dialogues rooted in $\langle \mathcal{A}_0, Q_0, \alpha_0\rangle$, formalized as a *dialectical tree*. [4]

**Definition 19 (Dialectical Tree)** *Let $\mathcal{P}$ be a DeLP program, and let $\langle \mathcal{A}_0, Q_0, \alpha_0\rangle$ be an argument wrt $\mathcal{P}$. A dialectical tree for $\langle \mathcal{A}_0, Q_0, \alpha_0\rangle$, denoted $\mathcal{T}_{\langle \mathcal{A}_0, Q_0, \alpha_0\rangle}$, is a tree structure defined as follows:*
*1) The root node of $\mathcal{T}_{\langle \mathcal{A}_0, Q_0, \alpha_0\rangle}$ is $\langle \mathcal{A}_0, Q_0, \alpha_0\rangle$.*
*2) $\langle \mathcal{B}', h', \beta'\rangle$ is an immediate child of $\langle \mathcal{B}, h, \beta\rangle$ iff there exists an acceptable argumentation line $\lambda^{\langle \mathcal{A}_1, Q_1, \alpha_1\rangle} = [\langle \mathcal{A}_0, Q_0, \alpha_0\rangle, \langle \mathcal{A}_1, Q_1, \alpha_1\rangle, \ldots, \langle \mathcal{A}_n, Q_n, \alpha_n\rangle, \ldots]$ such that there are two elements $\langle \mathcal{A}_{i+1}, Q_{i+1}, \alpha_{i+1}\rangle = \langle \mathcal{B}', h', \beta'\rangle$ and $\langle \mathcal{A}_i, Q_i, \alpha_i\rangle = \langle \mathcal{B}, h, \beta\rangle$, for some $i = 0 \ldots n-1$.*

**Example 20** *Consider $\langle \mathcal{A}_1, engine\_ok, 0.3\rangle$ from Example 12, and the argumentation line shown in Example 18. Note that the argument $\langle \mathcal{A}_2, \sim fuel\_ok, 0.6\rangle$ has a second (blocking) defeater $\langle \mathcal{A}_4, fuel\_ok, 0.6\rangle$, computed from (4), (16) via INTF and MPA, resp. The argument $\langle \mathcal{A}_1, engine\_ok, 0.3\rangle$ has also a second defeater $\langle \mathcal{A}_5, \sim engine\_ok, 0.95\rangle$, computed from (5), (11) via INTF and MPA, resp. There are no more arguments to consider. There are three acceptable argumentation lines rooted in $\langle \mathcal{A}_1, engine\_ok, 0.3\rangle$, namely:*

- $[\langle \mathcal{A}_1, engine\_ok, 0.3\rangle, \langle \mathcal{A}_2, \sim fuel\_ok, 0.6\rangle, \langle \mathcal{A}_3, \sim low\_speed, 0.8\rangle]$
- $[\langle \mathcal{A}_1, engine\_ok, 0.3\rangle, \langle \mathcal{A}_2, \sim fuel\_ok, 0.6\rangle, \langle \mathcal{A}_4, fuel\_ok, 0.9\rangle]$
- $[\langle \mathcal{A}_1, engine\_ok, 0.3\rangle, \langle \mathcal{A}_5, \sim engine\_ok, 0.95\rangle]$

*Fig. 2 shows the corresponding dialectical tree $\mathcal{T}_{\langle \mathcal{A}_1, engine\_ok, 0.3\rangle}$ rooted in $\langle \mathcal{A}_1, engine\_ok, 0.3\rangle$.*

Nodes in a dialectical tree $\mathcal{T}_{\langle \mathcal{A}_0, Q_0, \alpha_0\rangle}$ can be marked as *undefeated* and *defeated* nodes (U-nodes and D-nodes, resp.). A dialectical tree will be marked as an AND-OR tree: all leaves in $\mathcal{T}_{\langle \mathcal{A}_0, Q_0, \alpha_0\rangle}$ will be marked U-nodes (as they have no defeaters), and every inner node is to be marked as *D-node* iff it has at least one U-node as a child, and as *U-node* otherwise. An argument $\langle \mathcal{A}_0, Q_0, \alpha_0\rangle$ is ultimately accepted as valid (or *warranted*) wrt a DeLP program $\mathcal{P}$ iff the root of $\mathcal{T}_{\langle \mathcal{A}_0, Q_0, \alpha_0\rangle}$ is labelled as *U-node*.

**Example 21** *Consider the dialectical tree $\mathcal{T}_{\langle \mathcal{A}_1, engine\_ok, 0.3\rangle}$ from Example 20. The marking procedure results in the nodes of $\mathcal{T}_{\langle \mathcal{A}_1, engine\_ok, 0.3\rangle}$ marked as U-nodes and D-nodes as shown in Fig. 2.*[5]

**Definition 22 (Warrant)** *Given a program $\mathcal{P}$, and a goal $Q$, we will say that $Q$ is warranted wrt $\mathcal{P}$ with a necessity $\alpha$ iff there exists a warranted argument $\langle \mathcal{A}, Q, \alpha\rangle$.*

For a given program $\mathcal{P}$, a P-DeLP interpreter will find an answer for a goal $Q$ by determining whether $Q$ is supported by some warranted argument $\langle \mathcal{A}, Q, \alpha\rangle$. Different doxastic attitudes are distinguished when providing an answer for the goal $Q$ according to the associated status of warrant.
(1) Answer YES (with a necessity $\alpha$) whenever $Q$ is supported by a warranted argument $\langle \mathcal{A}, Q, \alpha\rangle$;
(2) Answer NO (with a necessity $\alpha$) whenever for $\sim Q$[6] is supported by a warranted argument $\langle \mathcal{A}, \sim Q, \alpha\rangle$;
(3) Answer UNDECIDED whenever (1) and (2) do not hold. It can be shown that the cases (1) and (2) cannot hold simultaneously [García and Simari, 2004]: if there exists an warranted argument for an atom $q$ based on a program $\mathcal{P}$, then there is no warranted argument for $\sim q$ based on $\mathcal{P}$.

**Example 23** *Consider program $\mathcal{P}_{eng}$, and the goal engine_ok. The only argument supporting engine_ok is not warranted (as shown in Fig. 2). On the contrary, there exists an argument $\langle \mathcal{A}_5, \sim engine\_ok, 0.95\rangle$ supporting $\sim engine\_ok$, and such argument has no defeaters, and therefore it is warranted. The answer to goal engine_ok will*

---

[4] It must be remarked that the definition of dialectical tree as well as the characterization of constraints to avoid fallacies in argumentation lines can be traced back to [Simari et al., 1994]. Similar formalizations were also used in other argumentation frameworks (e.g. [Prakken and Sartor, 1997]).

[5] The search space associated with dialectical trees is reduced by applying $\alpha-\beta$ pruning [García and Simari, 2004] (e.g. in Figure 2, if the right branch is computed first, then the left branch of the tree does not need to be computed).

[6] For a given goal $Q$, we write $\sim Q$ as an abbreviation to denote "$\sim q$" if $Q \equiv q$ and "$q$" if $Q \equiv \sim q$.

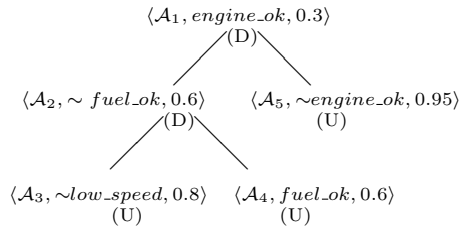

Figure 2: Dialectical tree for $\langle \mathcal{A}_1, engine\_ok, 0.3 \rangle$

*therefore be* No, *with* $\alpha = 0.95$.
*Consider now the same program* $\mathcal{P}_{eng}$, *and the goal* $fuel\_ok$. *The only argument supporting* $fuel\_ok$ *is* $\langle \mathcal{A}_4, fuel\_ok, 0.6 \rangle$, *which is defeated by a blocking defeater* $\langle \mathcal{A}_2, \sim fuel\_ok, 0.6 \rangle$. *The analysis for* $\langle \mathcal{A}_2, \sim fuel\_ok, 0.6 \rangle$ *is analogous, as this argument is defeated by* $\langle \mathcal{A}_4, fuel\_ok, 0.6 \rangle$. *Thus both arguments 'block' each other, neither of them being warranted. The resulting answer is* UNDECIDED.

## 5 Conclusions and related work

In this paper we have presented P-DeLP, a new logic programming language based on Defeasible Logic Programming which incorporates the treatment of possibilistic uncertainty and representation of fuzzy knowledge. The proposed approach improves the PGL logic programming framework, allowing to reason about conflicting goals on the basis of an argument-based procedure for computing warrant built on top of the PGL inference mechanism. In this approach, arguments are sets of uncertainty weighted formulas that support a goal, and support weights are used to resolve conflicts among contradictory goals. It must be remarked that DeLP has been successfully integrated in a number of real-world applications (e.g. clustering [Gómez and Chesñevar, 2004], intelligent web search [Chesñevar and Maguitman, 2004b] and recommender systems [Chesñevar and Maguitman, 2004a]). Several features leading to efficient implementations of DeLP have been also recently studied, particularly those related to comparing conflicting arguments by specificity [Stolzenburg et al., 2003] and defining transformation properties for DeLP programs [Chesñevar et al., 2003]. Extensions for DeLP in the context of multiagent systems have been also been proposed [Capobianco et al., 2004].

In this context, P-DeLP keeps all the original features of DeLP while incorporating more expressivity and representation capabilities by means of possibilistic uncertainty and fuzzy knowledge. One particularly interesting feature of P-DeLP is the possibility of defining aggregated preference criteria by combining the necessity measures associated with arguments with other syntax-based criteria (e.g. specificity [Simari and Loui, 1992, Stolzenburg et al., 2003]).

In the last years the development of combined approaches based on qualitative reasoning and uncertainty has deserved much research work [Parsons, 2001]. Preference-based approaches to argumentation have been developed, many of them oriented towards formalizing conflicting desires in multiagent systems [Amgoud, 2003, Amgoud and Cayrol, 2002]. In contrast with these preference-based approaches, the P-DeLP framework involves necessity measures explicitly attached to fuzzy formulas and the proof procedure of the underlying possibilistic fuzzy logic is used for computing the necessity measure for arguments. Besides, it must be stressed that a salient feature of P-DeLP is that it is based on two logical frameworks that have already been implemented (namely PGL [Alsinet and Godo, 2001] and DeLP [García and Simari, 2004]).

There has been generic approaches connecting defeasible reasoning and possibilistic logic (e.g.[Benferhat et al., 2002]), and recently a number of hybrid approaches connecting argumentation and uncertainty have been developed. Probabilistic Argumentation Systems [Haenni et al., 2000, Haenni and Lehmann, 2003] use probabilities to compute degrees of support and plausibility of goals, related to Dempster-Shafer belief and plausibility functions. However this is not a dialectics-based framework as opposed to the one presented in this paper. In [Schweimeier and Schroeder, 2001] a fuzzy argumentation system based on extended logic programming is proposed. In contrast with our framework, this approach relies only on fuzzy values applied to literals and there is no explicit treatment of possibilistic uncertainty.

Part of our current research work will be developed into three directions: first, we will extend the existing implementation of DeLP to incorporate the new features of P-DeLP. Second, we will apply the resulting implementation of P-DeLP to improve existing real-world applications of DeLP and to develop new ones. Finally, we will analyze extending P-DeLP to first order. It must be remarked that the Generalized Modus Ponens rule used in P-DeLP is syntactically the same as the one used in possibilistic logic [Dubois et al., 1994]. As a consequence, to implement the machinery of P-DeLP the underlying possibilistic fuzzy logic PGL can be replaced by the possibilistic logic. The advantage of this approach is that the current logic programming system can be extended to first order, incorporating fuzzy unification between fuzzy constants [Alsinet and Godo, 2001].

**Acknowledgements**

This research was partially supported by CYCYT Projects LOGFAC (TIC2001-1577-C03-01/03) and SOFTSAT (TIC2003-00950), by Ramón y Cajal Program (Ministerio de Ciencia y Tecnología, Spain), by CONICET (Argentina), by the Secretaría General de Ciencia y Tecnología de la Universidad Nacional del Sur and by Agencia Nacional de Promoción Científica y Tecnológica (PICT 2002 No. 13096). The authors would like to thank anonymous reviewers for providing helpful comments to improve the final version of this paper.